\documentclass[letterpaper, 10 pt, conference]{ieeeconf}  

\IEEEoverridecommandlockouts                              

\usepackage{dblfloatfix} 
\usepackage{graphicx}
\usepackage{url}
\usepackage{flushend}
\usepackage{cite}
\usepackage{subfigure}
\usepackage{amsmath,amsfonts,amssymb}
\usepackage{xcolor}
\usepackage{optidef}
\usepackage{algorithm}
\usepackage{algpseudocode}
\usepackage{multirow}
\usepackage{multicol}
\usepackage{hyperref}

\title{\LARGE
PUGS: \textbf{P}erceptual \textbf{U}ncertainty for \textbf{G}rasp \textbf{S}election in Underwater Environments
}

\author{Onur Bagoren$^{1}$, Marc Micatka$^{2}$, Katherine A. Skinner$^{1}$, and Aaron Marburg$^2$
\thanks{This work was funded by the Office of Naval Research under grants N00014-21-1-2052 and N00014-22-1-2196}
\thanks{$^1$O. Bagoren and K.A. Skinner are with the Department of Robotics, University of Michigan, Ann Arbor, MI 48109, USA}%
\thanks{$^2$M. Micatka and A. Marburg are with the Applied Physics Laboratory, University of Washington, Seattle, WA 98105, USA}%
\thanks{Corresponding author e-mail: {\tt\small obagoren@umich.edu}}%
}

\begin{document}

\maketitle
\thispagestyle{empty}
\pagestyle{empty}

\begin{abstract}
When navigating and interacting in challenging environments where sensory information is imperfect and incomplete, robots must make decisions that account for these shortcomings. 
We propose a novel method for quantifying and representing such perceptual uncertainty in 3D reconstruction through occupancy uncertainty estimation. 
We develop a framework to incorporate it into grasp selection for autonomous manipulation in underwater environments. 
Instead of treating each measurement equally when deciding which location to grasp from, we present a framework that propagates uncertainty inherent in the multi-view reconstruction process into the grasp selection.
We evaluate our method with both simulated and the real world data, showing that by accounting for uncertainty, the grasp selection becomes robust against partial and noisy measurements. 
Code will be made available at \url{https://onurbagoren.github.io/PUGS/}
\end{abstract}
\section{Introduction}
Underwater remotely operated vehicles (ROVs) equipped with manipulator systems allow for advancements in ocean cleaning operations \cite{guimond_finding_plastic_2024}, deep sea specimen sampling \cite{mazzeo_marine_2022}, and human-robot collaboration in underwater environments \cite{feng_overview_2020}. 
Standard underwater manipulator systems are often controlled through teleoperation \cite{sivcev_underwater_2018}. 
Despite its success, reliance on teleoperation poses many challenges, primarily in operational cost and pilot training time.
These challenges associated with teleoperation have motivated research advances in autonomous underwater manipulation \cite{petillot_underwater_2019,micatka_grasp_volumes_2024}, where the operational cost is significantly lower.
Yet, many challenges remain in deploying autonomous mobile manipulator systems in environments such as the underwater domain.

Providing informative representations of the environment and objects that robots interact with is critical to achieving safe and reliable autonomy for robotic systems.
For this, a common approach is using sensors that provide rich environmental information to map the surrounding scene \cite{wang_real-time_2023, chen_monorun_2021, song2024turtlmap}. 
However, underwater sensing and perception suffer from several challenges, including image degradation \cite{mobley_light_1994, li_watergan_2017}, sensor noise \cite{song_uncertainty-aware_2023}, and uncertainty due to a lack of absolute position measurements \cite{rahman_svin2_2022}, which lead to challenges for robotics tasks that rely on accurate perceptual information.

A growing number of works have been successful in showing that incorporating uncertainty into perception systems, whether for detection \cite{kendall-what-uncertainties}, simultaneous localization and mapping~\cite{song_uncertainty-aware_2023}, or 3D reconstruction \cite{rosinol_probabilistic_2022,goli2023, ulusoy_patches_2016}, plays a crucial role in improving performance.
In addition to playing a crucial role in improving the robustness of perception systems, uncertainty plays a key role in embodied robotic tasks, such as planning and manipulation \cite{pmlr-v164-saund22a}. 

In this work, we model perceptual uncertainty in 3D reconstruction tasks to improve grasp selection and the manipulation of objects in underwater environments, using a method we call \textbf{P}erceptual \textbf{U}ncertainty for \textbf{G}rasp \textbf{S}election (PUGS). 
\begin{figure}[t!]
    \centering
    \includegraphics[width=1.0\linewidth]{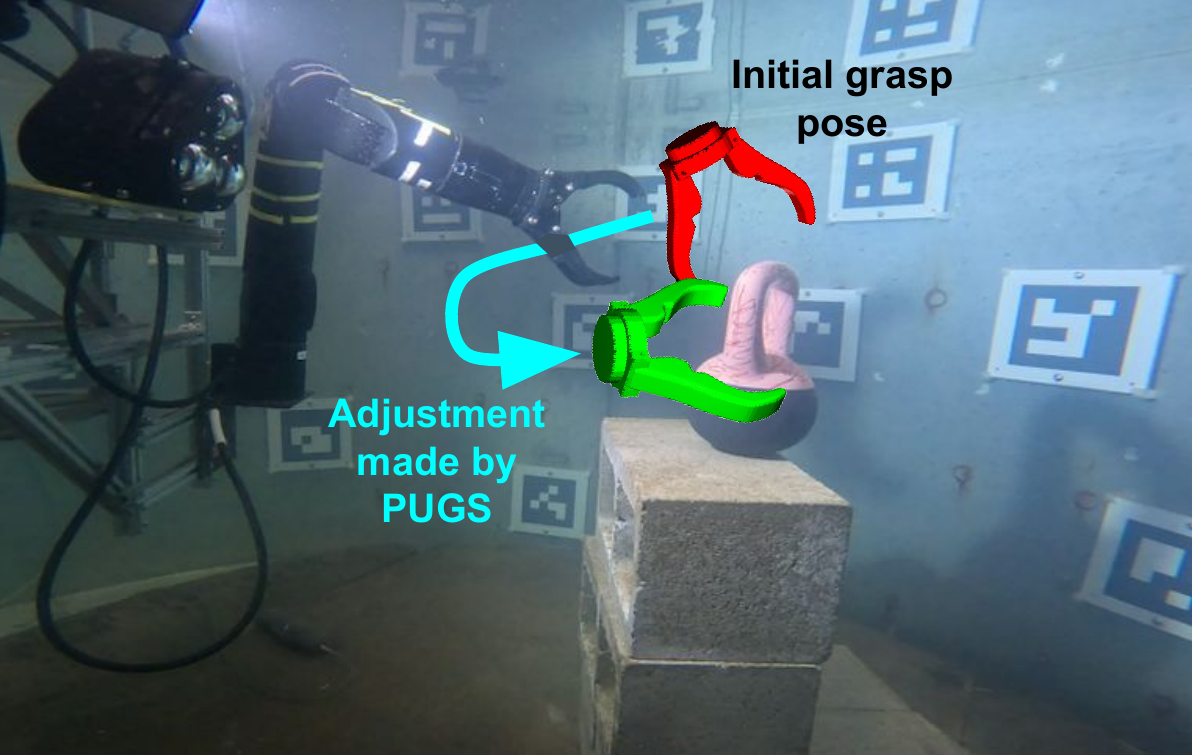}
    \caption{Real-world underwater manipulation setup in the test tank at the University of Washington Applied Physics Laboratory (UW-APL). 
    The test system includes a Reach Robotics Bravo 7 electric manipulator and Trisect subsea stereo sensor.
    The gripper in red shows the grasp pose from an out-of-the-box grasp selection model \cite{player_real-time_2023}.
    The proposed method successfully leads to a more reliable grasping location, shown in green, by using perceptual uncertainty to weigh more favorable grasp locations.
    The visualizations are taken from evaluations of a real-world dataset and overlaid to match the photo showing the robot during regular operation.
    }
    \label{fig:hh101}
    \vspace{-3mm}
\end{figure}
We focus on modeling how uncertainty inherent from multi-view stereo can be leveraged for quantifying uncertainty in 3D reconstruction, specifically in representing occupancy in 3D space.
We then show that the uncertainty of the occupied regions can be a useful for improving existing grasp selection methods and guiding toward more reliable and robust grasp selection. We present the following contributions:
\begin{enumerate}
    \item We propose the construction of a fused occupancy field (FOF) informed by the uncertainty measurements and pose estimates. 
    \item We develop a novel method to quantify the predictive uncertainty associated with occupancy in 3D space using probabilistic regression methods.
    \item We present an uncertainty fusion method to combine information from measurement and predictive uncertainty for modeling occupancy uncertainty.
    \item We provide an experimental evaluation in both simulation and real-world underwater environments to validate the proposed methods.
\end{enumerate}

\section{Related Works}
\begin{figure*}[t!]
    \centering
    \includegraphics[width=1.0\textwidth]{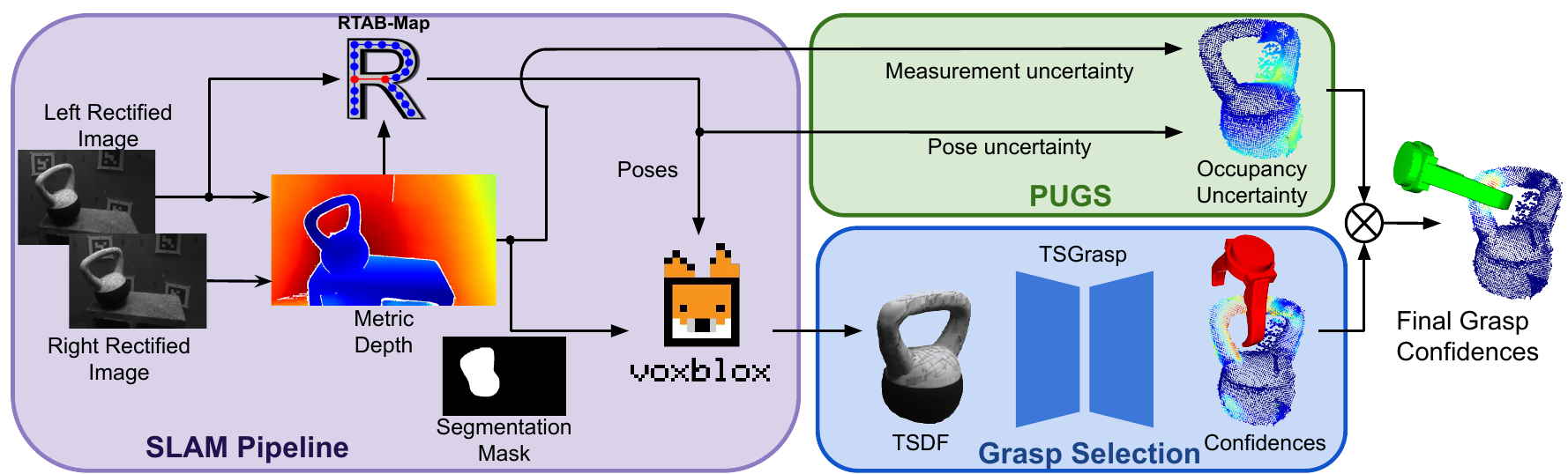}
    \caption{Overview of the proposed system. We use a SLAM pipeline to construct dense 3D reconstructions of objects of interest.
    PUGS quantifies the uncertainty inherent in the observation and pose estimation to construct an occupancy uncertainty representation. We use TSGrasp \cite{player_real-time_2023} as the baseline network to regress on grasp poses and confidences.
    The final grasp confidence and pose are determined by fusing the PUGS occupancy uncertainty output and the TSGrasp confidences.}
    \label{fig:overview}
\end{figure*}
\subsection{Underwater Manipulation and Grasp Selection}
Compared to traditional robotic manipulation, manipulation with underwater vehicle manipulation systems (UVMS) has additional challenges associated with vehicle dynamics, sensing, and environmental disturbances \cite{billings_hybrid_2022, chang_adaptive_nodate, player_real-time_2023}.

For improving the perception in underwater environments, Billings et al. present a hybrid perception system for a UVMS with a stereo rig mounted on the vehicle and a fish eye camera mounted on the manipulator, with an accompanying mapping and pose estimation method\cite{billings_hybrid_2022}.
Chang et al. present an information-theoretic approach to the underwater planning and grasping task, where the information associated with the object pose and shape is aimed to be maximized \cite{chang_adaptive_nodate}.
Different from \cite{billings_hybrid_2022}, our proposed system aims to use robust perception for grasping rather than pose estimation and mapping, where an improvement of the perception and grasping capabilities are achieved through an uncertainty-centric approach. 

TSGrasp, a grasp selection method tested in an underwater setting with a similar hardware setup, builds on previous work \cite{mousavian_6-dof_2019} by utilizing sparse spatiotemporal convolution to output grasp poses, gripper widths, and confidences on an input pointcloud \cite{player_real-time_2023}.
This method is tested on point clouds accumulated from 1 or 4 frames of images and utilizes geometric information only for the grasp selection.
PUGS is a method that builds upon TSGrasp's grasping capabilities by formulating a fusion between grasp confidence and occupancy uncertainty.
For this reason, along with the fact that it was developed for underwater grasp selection, we select TSGrasp as a baseline method to evaluate PUGS and show that our proposed method, accounting for uncertainty, improves grasping capabilities in adversarial situations.


\subsection{Uncertainty Representation for 3D Reconstruction}
Uncertainty representations in 3D reconstruction aim to capture the uncertainty originating from the measurement model or multi-view stereo to improve the reconstruction quality \cite{freundlich_exact_2015, ulusoy_towards_2015, ulusoy_patches_2016, rosinol_probabilistic_2022, liao_multi-view_2024}, as opposed to classical methods that treat each measurement equally in the reconstruction process \cite{labbe2019rtab}.

Freundlich et al. compute the bias and covariance associated with 3D locations obtained from stereo imaging, showing that the calculated bias correction leads to more accurate representations\cite{freundlich_exact_2015}.
Ulusoy et al. present a series of papers on using probabilistic methods for volumetric rendering~\cite{ulusoy_towards_2015,ulusoy_patches_2016}.
The developed framework is compared against classical methods that perform point estimates on occupancy and show that the relation between appearance and occupancy are captured more accurately through a probabilistic approach.
Rosinol et al. show that estimating the uncertainty from depth estimation and using it as a weighing factor for volumetric fusion results in fast and accurate reconstruction with little ad-hoc filtering \cite{rosinol_probabilistic_2022}.
Learning-based methods have also shown promise in improving reconstruction quality through jointly learning the uncertainty associated with multi-view reconstruction \cite{liao_multi-view_2024}.
We formulate PUGS in a similar vein.
Rather than using uncertainty as a prior for improving reconstruction quality, we present PUGS as a method to improve the robustness of grasp selection.

Methods that utilize uncertainty-aware reconstruction have also been used for downstream robotics tasks in various applications \cite{von_drigalski_uncertainty-aware_2022, torroba_fully-probabilistic_2022, 10093134, pmlr-v164-saund22a}.
Relevant to the manipulation task, Von Drigalski et al. proposed a framework for representing the uncertainty of the pose of an object before and after being manipulated. 
The uncertainty representation enabled accurate and efficient manipulation through accurate belief propagation \cite{von_drigalski_uncertainty-aware_2022}.
Torroba et al. show that accounting for the uncertainty of an estimated 2.5D height map produced from dense range measurements helps improve the localization capabilities of subsea mobile robotic platforms \cite{torroba_fully-probabilistic_2022, 10093134}.
Saund et al. propose a grasp selection method that uses RGB-D and tactile information to manipulate the object's shape, with uncertainty propagated from a learned latent representation to inform the end reconstruction~\cite{pmlr-v164-saund22a}.
We select our uncertainty representation to be associated with the occupancy measured, as opposed to the pose of an object \cite{von_drigalski_uncertainty-aware_2022}, the shape of the object~\cite{pmlr-v164-saund22a} or a height map~\cite{torroba_fully-probabilistic_2022, 10093134}.
This uncertainty is then used directly to improve the robustness and autonomy of a robotic task in grasp selection.

\section{Methodology}
The complete pipeline for PUGS is shown in Fig. \ref{fig:overview}.
We take in rectified stereo image pairs as input and estimate camera pose and depth measurements within our SLAM pipeline.
We quantify the uncertainties associated with depth and pose estimates, and integrate them to compute occupancy uncertainty. 
We use occupancy uncertainty as a weighting factor for confidence output in grasp selection.



We consider a scenario in which we have an object of interest to grasp and a set of camera views in which this object is visible. 
Going forward, we assume we have the following systems: a vision-based SLAM system for pose estimation~\cite{labbe_rtabmap_2019}, a semantic segmentation model for segmenting the object of interest in the image~\cite{wu2019detectron2}, and depth estimates computed from the calibrated stereo rig~\cite{lipson2021raft}.
We describe how, given the measurements and systems in place, an uncertainty representation for grasping can be formulated and used for reliable robotic manipulation in adversarial environments such as underwater environments.

\subsection{Multi-view Uncertainty in 3D Representations}
We aim to model the uncertainty associated with occupancy based on observations from a multi-view camera system.
For this, we consider two primary modes of uncertainty that play a role in reconstruction and grasp selection: observational uncertainty and predictive uncertainty. 

\subsubsection{Notation and Formal Definitions}
We use bold variables $\boldsymbol{x}$ to represent a vector, and non-bolded variables $x$ to represent scalar values.
Random variables and vectors are notated as $\widehat{x}$.

From the given set-up, we compute depth images $\mathbf{D} = \{\widehat{D}^i\}_{i=1}^N$ from each stereo image pair. 
We make the modeling assumption that each pixel $\widehat{d}_{\text{uv}}^i$ in the depth image $\widehat{D}^i$ can be modeled as a Gaussian random variable $\widehat{d}_{\text{uv}}^i \sim \mathcal{N}(\mu_\text{uv}^{(i)}, \sigma^{2(i)}_\text{uv})$, where $\mu_\text{uv}^{(i)}$ is the depth measurement at pixel $(u,v)$ for frame $i$ and $\sigma^{2(i)}_{\text{uv}}$ is the associated measurement noise.
The noise $\sigma^{2(i)}_{\text{uv}}$ can be obtained from either the sensor specifications or the model used to compute the depth from the stereo images. 

From the vSLAM system, we obtain a set of poses $\mathbf{C} = \{\widehat{C}^i\}_{i=1}^N$ synchronized with each stereo pair RGB image and depth image.
We assume that the estimated poses are modeled as Gaussian random variables such that $\widehat{C}_i \sim \mathcal{N}(\boldsymbol{\mu}_C^i, \mathbf{\Sigma}_C^i)$, where $\boldsymbol{\mu}_C \in SE(3)$ is the estimated pose of the camera at frame $i$ and $\mathbf{\Sigma_C^i}$ is the associated pose covariance.

\subsection{Observational Uncertainty Representation}\label{ss:observational_unc}
Observational uncertainty is modeled using the measurement noise and the pose uncertainty.
Based on these estimates, these two factors are propagated into a fused occupancy field (FOF) to model areas that are likely to be occupied.

\subsubsection{Measurement Noise Propagation} \label{sss:mnp}
Since each measurement used to compute occupancy has inherent uncertainty, we aim to model how this uncertainty can be estimated and used for occupancy computation.
In our setup, we consider occupancy to be measured by the depth estimate.
We backproject pixel $(u,v)$ into the camera frame as $\widehat{\mathbf{p}}_c \sim \mathcal{N}(\boldsymbol{\mu}_{\mathbf{p}_c}, \mathbf{\Sigma}_{\mathbf{p}_c})$ using the uncertain depth measurement $\widehat{d}_{uv}^i$. The mean $\boldsymbol{\mu}_{\mathbf{p}_c} \in \mathbb{R}^3$ is the position in the camera frame and $\mathbf{\Sigma}_{\mathbf{p}_c}$ the positional covariance. 
The backprojection model is shown in Eq. \eqref{eq:unproj_mean_cov}, where $\left(f_x, f_y, c_x, c_y\right)$ are the pinhole camera intrinsic parameters.
To compute the covariance, we take the first-order linearization of the backprojection model with respect to the depth variable and propagate the covariance originating from the depth variance into the camera frame using the Jacobian $\boldsymbol{J}_d$ in Eq. \eqref{eq:jacobian}.
\begin{gather}
    \boldsymbol{\mu}_{\mathbf{p}_c} = \mu_\text{uv}^{(i)}\begin{bmatrix}\frac{u-c_x}{f_x} \\ \frac{v-c_y}{f_y} \\ 1\end{bmatrix}, \ \  
    \mathbf{\Sigma}_{\mathbf{p}_c} = \boldsymbol{J}_d \sigma_\text{uv}^{2(i)} \boldsymbol{J}_d^T \label{eq:unproj_mean_cov} \\
    \boldsymbol{J}_d = \frac{\partial{\mathbf{p}_c}}{\partial d_\text{uv}} = \begin{bmatrix}\frac{u-c_x}{f_x} & \frac{v-c_y}{f_y} & 1\end{bmatrix}^T \label{eq:jacobian}
\end{gather}

\subsubsection{Pose Uncertainty Propagation} \label{sss:pup}
The estimated poses are inherently uncertain due to accumulating errors and the absence of absolute position measurements.
The uncertainty of poses when measurements are made contributes to uncertainty in computed occupancy.
To model this, we use a pose uncertainty propagation mechanism.

We place $\widehat{\mathbf{p}}_c$ from the camera into the world frame as $\widehat{\mathbf{p}}_w \sim \mathcal{N}(\boldsymbol{\mu}_{\mathbf{p}_w}, \mathbf{\Sigma}_{\mathbf{p}_w})$ using the pose estimate $\widehat{C}^i$ as shown Eq. \eqref{eq:cam_to_world}. 
The mean $\boldsymbol{\mu}_{\mathbf{p}_w}$ is the position in the world coordinate frame and $\mathbf{\Sigma}_{\mathbf{p}_w}$ the positional covariance.
Here, $\mathbf{\Sigma}_{C_t}^i$ is the covariance associated with the translation, $\text{R}_c^w \in SO(3)$ is the rotation and $\text{t}_c^w \in \mathbb{R}^3$ is the translation component of $\widehat{C}^i$.
\begin{gather}
    \boldsymbol{\mu}_{\mathbf{p}_w} = \text{R}_c^w \boldsymbol{\mu}_{\mathbf{p}_c} + \text{t}_c^w,\ \ \mathbf{\Sigma}_{\mathbf{p}_w} = \mathbf{\Sigma}_{\mathbf{p}_c} + \mathbf{\Sigma}_{C_t}^i \label{eq:cam_to_world}
\end{gather}

\subsubsection{Fused Occupancy Field} \label{sss:FOFc}
Through the FOF, we aim to represent the occupancy field based on the measurements and their uncertainties.
We assume that the probability distribution of $\widehat{\mathbf{p}}_w$ can be used as an informative representation of the occupancy density in the 3D space in which it exists.

For each of the $\mathbf{N}$ points in the world frame $\widehat{\mathbf{p}}_w^{(i)}$, we take a weighted sum of the probability distribution functions to obtain a Gaussian Mixture Model (GMM) representation of the occupancy field, as shown in Eq. \eqref{eq:gmm}.
This modeling choice for the occupancy field allows us to maintain a continuous and generative representation \cite{gmm_perception}, which we can then query to compute the occupancy of new measurements.
\begin{gather}
    \text{OF}(\mathbf{p}) = \frac{1}{\mathbf{N}}\sum_{j=1}^{\mathbf{N}} \mathcal{N}(\mathbf{p} \lvert \boldsymbol{\mu}_{\mathbf{p}_w}^{(j)}, \mathbf{\Sigma}_{\mathbf{p}_w}^{(j)}) \label{eq:gmm}
\end{gather}

Although Eq. \eqref{eq:gmm} is sufficient for representing the occupancy field, it neglects the capability to fuse measurements that share information.
To address this, we add an extra measurement fusion mechanism.
For a new measurement $\mathbf{p}_z \in \mathbb{R}^3$, we find the $K$ nearest Gaussians $\widehat{\mathbf{P}}_K = \{\widehat{\mathbf{p}}_w^i\}_{i=1}^K$. 
We then perform a weighted Bayesian fusion \cite{barfoot_state_2017} to compute the resulting Gaussian distribution of $\mathbf{p}_z$, as shown in Eq. \eqref{eq:bayesian_fusion}.
Each weight for the $K$ nearby Gaussians is determined by the responsibility of Gaussian $\widehat{\mathbf{p}}_w^i$ for $\mathbf{p}_z$, shown in Eq. \eqref{eq:responsibility}, where $\pi_k=\frac{1}{K}$.
\begin{gather}
    \mathbf{\Sigma}_{\mathbf{p}_z} = \left(\sum_{k=1}^K\gamma_k(\mathbf{p}_z)\mathbf{\Sigma}_k^{-1}\right)^{-1} \label{eq:bayesian_fusion}\\
    \gamma_k(\mathbf{x}) = \frac{\pi_k \mathcal{N}(\mathbf{x} | \boldsymbol{\mu}_k, \boldsymbol{\Sigma}_k)}{\sum_{j=1}^{K} \pi_j \mathcal{N}(\mathbf{x} | \boldsymbol{\mu}_j, \boldsymbol{\Sigma}_j)},\ \ \sum_{j=1}^K\pi_j = 1 \label{eq:responsibility}
\end{gather}
When we take the query point as the mean of each unprojected Gaussian and compute the occupancy field as shown in Eq. \eqref{eq:gmm}, we end up with a fused occupancy field (FOF).
This modeling allows us to represent the positional uncertainty of a new observation given past observations and the occupancy of a region based on depth estimates.
\subsection{Predictive Uncertainty Representation} \label{ss:pred_unc}
We model the predictive uncertainty to capture phenomena such as the frequency at which a certain region has been observed.
Intuitively, the aim is to obtain a representation that yields lower variance in regions with higher observation frequencies.
To model this, we used a stochastic variational Gaussian process (SVGP) \cite{hensman_scalable_nodate} to regress on the FOF, as the FOF provides an informative representation of occupancy density.
We select SVGPs over standard Gaussian Processes as SVGPs have been shown to be capable of running real-time, dense probabilistic mapping \cite{torroba_fully-probabilistic_2022}.

By performing a probabilistic regression on the FOF, we can obtain the variance associated with the occupancy at each point in space, allowing us to quantify the uncertainty of the occupancy.


\subsubsection{Predictive Occupancy Uncertainty Estimation}
To train the SVGP, we use the FOF, which is constructed from each of the backprojected depth points $\mathbf{P}_w = \{\mathbf{p}_w^i\}_{i=1}^{\mathbf{N}}$.
After the training has converged, we query the SVGP model with a set of $Q$ query points $\mathbf{P}_Z = \{\mathbf{p}_Z^i\}_{i=1}^Q$. 
\begin{gather}
    \text{SVGP}(\mathbf{P}_Z) = \{\mu_Z^{(i)}, \sigma_Z^{2(i)}\}_{i=1}^Q \label{eq:gp} \\
    \sigma_{\text{pred}}^2 = \frac{\sigma_Z^{2(i)}}{1 + \lvert\mu_Z^{(i)}\rvert} \label{eq:pred_uncertainty}
\end{gather}

\noindent In Eq. \eqref{eq:gp}, mean $\mu_Z^{(i)}$ represents the occupancy density of point $\mathbf{p}_Z^{(i)}$, while $\sigma_Z^{2(i)}$ represents the variance of the occuancy density.
We use the two outputs in order to obtain the final predictive uncertainty as shown in Eq. \eqref{eq:pred_uncertainty}, which scales the variance by the density of the point $\mathbf{p}_Z^{(i)}$.

We elect to follow this modeling choice as the variance of the SVGP regression is dictated by the spatial frequency on the input domain rather than the regressed function value.
Through Eq. \eqref{eq:pred_uncertainty}, we aim to mitigate the fact that predictive variance should be dictated not just by spatial frequency but also by occupancy density.

\subsection{Uncertainty Fusion}\label{ss:unc_fusion}
In Sec. \ref{ss:observational_unc}, we describe how to obtain positional uncertainty of points in the world frame from depth measurements.
The trained SVGP models the predictive occupancy uncertainty at these points.
We propose a fusion method between these two representations to create a final \textbf{occupancy uncertainty} representation.

We take inspiration from Torroba et al. \cite{torroba_fully-probabilistic_2022} in performing this fusion through cubature integration and sigma point propagation.
Cubature integration allows us to use the shape of the covariance associated with the positional uncertainty.
To perform cubature integration, we determine integration points, $\mathbf{p}_n$, and mean and variance integration weights, $\boldsymbol{w}_\mu, \boldsymbol{w}_{\sigma^2}$, respectively.
The number of integration points is determined by the dimension $d$ of the space in which integration occurs ($d=3$ for 3D), and the integration points and weights are determined using the cubature spread parameters $\left(\alpha, \beta, \kappa\right)$, as shown in Eqs. \eqref{eq:lambda}-\eqref{eq:cubature}.

\begin{gather}
 \lambda = \alpha^2\left(d + \kappa\right)-d, \ \ \hfill \boldsymbol{p}_n^{(i)} = u^{(i)}\sqrt{d+\lambda} \label{eq:lambda}\\
    u = \begin{Bmatrix}
       \begin{bmatrix}
           0 \\ 0 \\  0
       \end{bmatrix}, & \begin{bmatrix}
           \pm1 \\ 0 \\  0
       \end{bmatrix}, & \begin{bmatrix}
           0 \\ \pm1 \\  0
       \end{bmatrix},
       & \cdots
    \end{Bmatrix},\ \vert u \rvert = 2d+1 \\
    \boldsymbol{w}^{(i)}_\mu = \begin{cases}
        \frac{\lambda}{d + \lambda},& \text{i}=0 \\
        \frac{\lambda}{2(d + \lambda)}, & \text{i}=1,2,\dots,2d
    \end{cases} \\
    \boldsymbol{w}^{(i)}_{\sigma^2} = \begin{cases}
        \frac{\lambda}{d + \lambda} + \left(1 - \alpha^2 + \beta\right),& \text{i}=0 \\
        \frac{\lambda}{2(d + \lambda)}, & \text{i}=1,2,\dots,2d
    \end{cases}
    \label{eq:cubature}    
\end{gather}

The integration points $\{\boldsymbol{p}_n^{(i)}\}_{i=1}^{2d+1}$ are used to query the trained SVGP to obtain the occupancy mean and predictive variance at each point.
We then perform cubature integration, as shown in Alg. \ref{alg:fusion}.
The output of this integration yields the final occupancy variance, $\sigma_{occ}^2$, for a query point, $p_z$.
\begin{table*}[b!]
\centering
\caption{Grasp success evaluation in the simulation environment. We run $N=5$ experiments for each test setup and report the percent grasp success by counting the number of successful attempts against the total number of attempts.}
\label{tab:quant_res_counting}
\begin{tabular}{c|l|cc|cc|cc}
 &
   &
  \multicolumn{2}{c|}{Grasp Traversal Success $\uparrow$} &
  \multicolumn{2}{c|}{Gripper Closed Onto Object $\uparrow$} &
  \multicolumn{2}{c}{Goal Pose Reached $\uparrow$} \\ \hline \hline
Method &
  Reconstruction Type &
  Kettlebell &
  Coffee Mug &
  Kettlebell &
  Coffee Mug &
  Kettlebell &
  Coffee Mug \\ \hline
TSGrasp \cite{player_real-time_2023} & Partial & 0\%          & 40\%          & 0\%          & 0\%          & 0\%          & 0\%          \\
PUGS (ours)                          & Partial & \textbf{60\%} & \textbf{60\%} & \textbf{60\%} & \textbf{40\%} & \textbf{60\%} & \textbf{40\%} \\ \hline
TSGrasp \cite{player_real-time_2023} & Noisy Partial            & 0\%          & 0\%          & 0\%          & 0\%          & 0\%          & 0\%          \\
PUGS (ours)                          & Noisy Partial            & \textbf{40\%}          & \textbf{80\%}            & 0\%          & \textbf{60\%}          & 0\%          & \textbf{60\%}          \\ \hline
TSGrasp \cite{player_real-time_2023} & Complete    & \textbf{100\%} & 60\%          & \textbf{100\%} & \textbf{80\%} & \textbf{100\%} & \textbf{60\%} \\
PUGS (ours)                          & Complete    & \textbf{100\%} & \textbf{80\%} & 60\%          & 60\%          & 60\%          & 20\%         
\end{tabular}
\end{table*}

\begin{algorithm}
\caption{Uncertainty and Predictive Uncertainty Fusion}
\begin{algorithmic}[1]
\Require Query point $p_z$
\State $\mathbf{P}_k \gets \texttt{KDTree}(\mathbf{P}_w).\texttt{query}(p_z)$
\State $\mathbf{\Sigma}_{\mathbf{p}_z} \gets \texttt{bayesianFusion}(p_z, \mathbf{P}_k)$ \Comment{Eq. \eqref{eq:bayesian_fusion}}
\State $\mathbf{p}_n, \boldsymbol{w}_\mu, \boldsymbol{w}_{\sigma^2} \gets \texttt{cubature}(p_z, \mathbf{\Sigma}_{\mathbf{p}_z})$ \Comment{Eqs. \eqref{eq:lambda}-\eqref{eq:cubature}}
\State $\boldsymbol{\mu}_n, \boldsymbol{\sigma}_\text{pred}^2 \gets \texttt{SVGP}\left(\mathbf{p}_n\right)$ \Comment{Sec. \ref{ss:pred_unc}}
\State $\mu_\text{occ} \gets \sum_i \boldsymbol{w}_\mu^{(i)} \boldsymbol{\mu}_n^{(i)}$
\State $\sigma^2_\text{occ} \gets \sum_i \boldsymbol{w}_{\sigma^2}^{(i)} \boldsymbol{\sigma}_\text{pred}^{2(i)}$
\State \Return $\{\mu_\text{occ}, \sigma_\text{occ}^2\}$
\end{algorithmic}
\label{alg:fusion}
\end{algorithm}


\subsection{Fusing Occupancy Uncertainty with Grasp Selection}\label{ss:grasp_selection}
For the grasp selection, we use the computed occupancy uncertainty with the outputs of a pre-trained grasp selection framework, such as TSGrasp \cite{player_real-time_2023}.
This allows us to balance the geometric reasoning inherent in the grasp selection method with the uncertainty over the constructed geometry.
This weighing is shown in Eq. \eqref{eq:grasp_weighing}, where $\left(\sigma_\text{occ}^2\right)_G$ is the occupancy variance for each point in $\mathbf{p}_G$, $GS\left(\cdot\right)$ a grasp selection method that outputs grasp confidences given a point, and $\nu$ the scaling parameter to control how much we want the uncertainty to be weighted.
\begin{gather}
    \tilde{\mathbf{c}}_G = \frac{GS\left(\mathbf{p}_G\right)}{\left(\sigma_\text{occ}^2\right)_G^\nu} \label{eq:grasp_weighing}
\end{gather}

The grasp pose is selected to be the pose associated with the highest confidence computed in Eq. \eqref{eq:grasp_weighing}.


\section{Experimental Setup}
\subsection{Implementation Details}
We treat the depth noise as static across frames and fix it as $\sigma_{uv}^2 = 0.001$ m.
We downsample input images by a factor of 4 to improve the run time of the uncertainty computation.
We use GPyTorch \cite{gardner2018gpytorch} to train the SVGP with a learning rate of $1e-3$ and $500$ inducing points over 100 epochs.
We use TS-Grasp \cite{player_real-time_2023} for the grasp selection algorithm in real-world and simulation experiments.
Throughout all experiments, we set the weight factor hyperparameter used in Eq. \eqref{eq:grasp_weighing} to $\nu = 5$.

The depth measurements at each pose are converted to point clouds to construct a fused mesh at 3 mm resolution using voxblox \cite{oleynikova2017voxblox}.
The output mesh is sampled to obtain a fused pointcloud, filtered by removing statistical outliers outside a threshold standard deviation ratio of $\sigma_\text{thresh}=0.01$, using Open3D \cite{Zhou2018}.
The filtered pointcloud is input into TSGrasp \cite{player_real-time_2023} and PUGS to obtain the final grasp weight confidences.

\begin{figure}[t]
    \centering
    \includegraphics[width=0.9\linewidth]{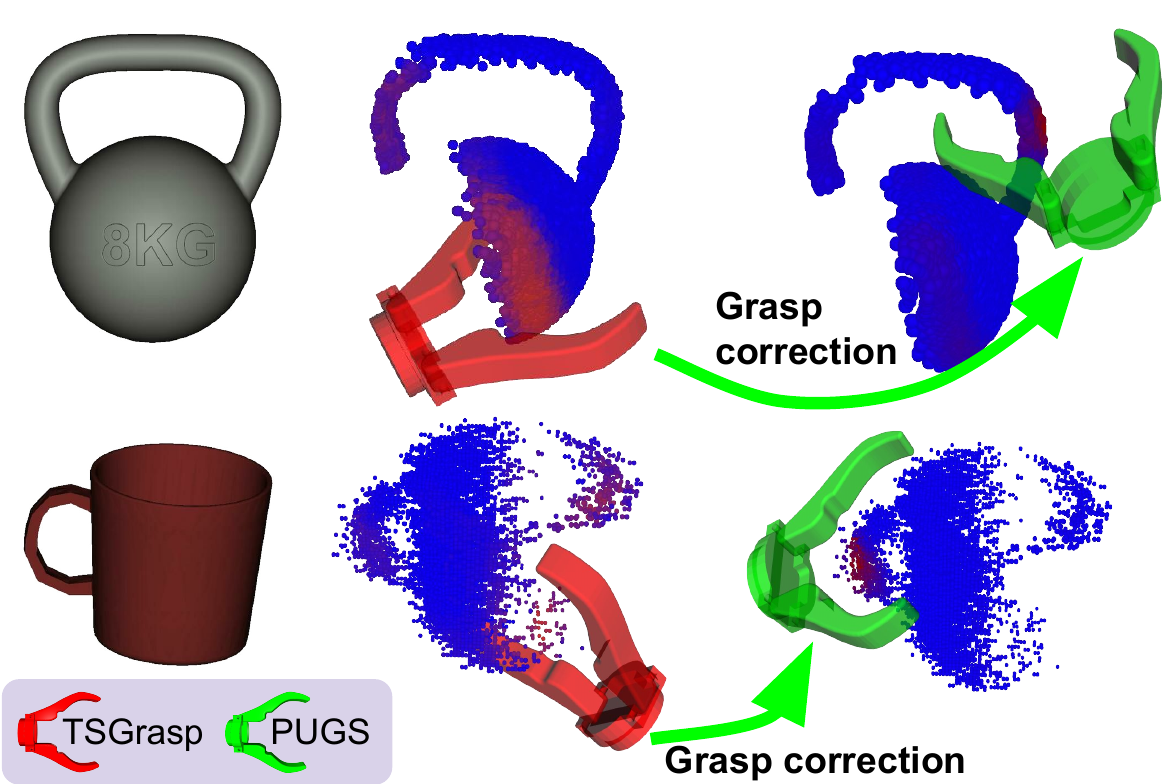}
    \caption{Selected results from simulation experiments with partial reconstructions of the kettlebell (first row) and noisy partial reconstructions of the coffee mug (second row). The first column shows the 3D object; the second column shows the grasps proposed by TSGrasp \cite{player_real-time_2023}, and the third column shows the grasps after the adjustment made by PUGS.
    The first row shows the partially reconstructed kettlebell and the ability of PUGS to recover from an incorrect grasp pose prediction where TSGrasp attempts to grasp from the edge of the partial reconstruction and PUGS leads to a more reliable grasp.
    The second row shows a noisy reconstruction of the coffee mug, and a corrected grasp prediction by PUGS.}
    \label{fig:sideways_fail}
\end{figure}

\begin{figure*}[t]
    \centering
    \includegraphics[width=0.85\textwidth]{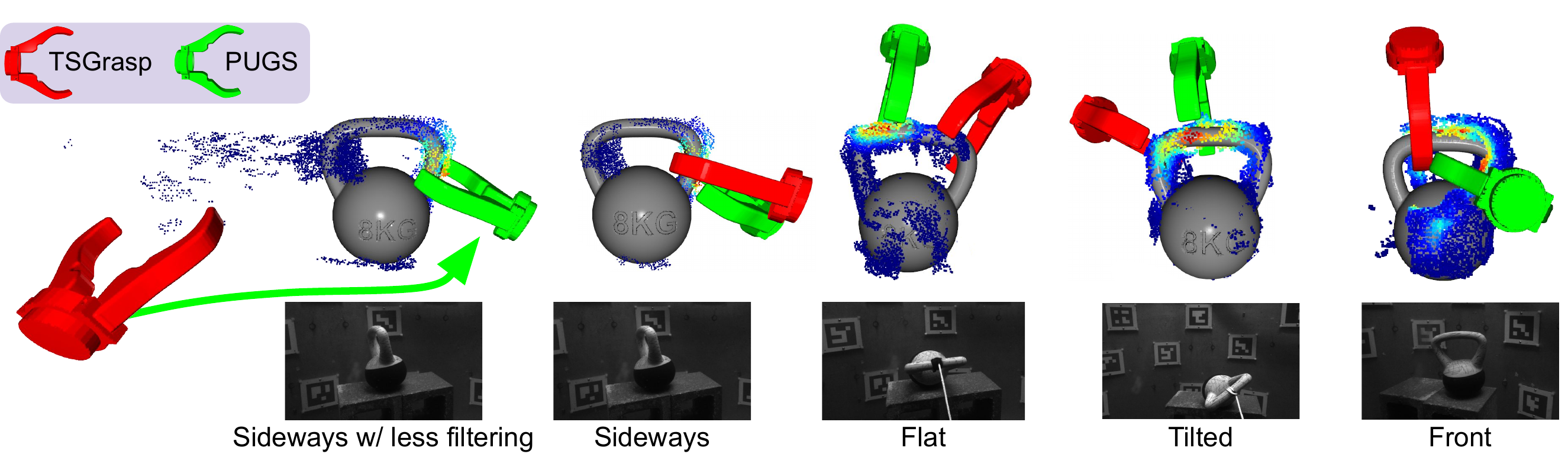}
    \caption{Results from data collected from the test tank. 
    The columns represent separate logs collected in the test tank.
    Each log contains images of partial views from different angles of the kettlebell.
    The first and second columns are from the same log, but the first column shows the pointcloud without the aggressive filtering.
    Here, we show a qualitative comparison of the proposed grasp poses from PUGS (green) and TSGrasp (red) \cite{player_real-time_2023}.
    The reconstructions are colored by the weighted confidence output of PUGS and overlaid on a representative 3D model of the object for easy visualization.
    Higher confidence regions are colored in red, and lower confidence regions are colored in blue.
    }
    \label{fig:qualitative_results}
    \vspace{-4mm}
\end{figure*}

\subsection{Simulation Setup}
We run experiments on a kettlebell and a coffee mug, shown in Fig. \ref{fig:qualitative_results}. 
For the simulation, we take a 3D model of the object and simulate object-centric camera poses.
We use Blender \cite{blender} to obtain the RGB and depth images.
We mask out the object in the rendered images to ensure that the grasp detection reasons only over the object.

Each RGB and depth image, camera pose, and mask are saved into a \texttt{rosbag} \cite{ros}, which we use to play back the simulated setup in a Gazebo \cite{gazebo} environment containing a Bravo manipulator \cite{noauthor_underwater_nodate} model.

We evaluate three types of reconstruction for each object: complete, partial, and noisy partial.
We set the depth noise for the noisy reconstruction to $\sigma_{uv}^2=0.01$ m and $\sigma_{uv}^2=0.001$ m for the other experiments.
When running the experiments, we note that the grasp pose proposals from PUGS and TSGrasp \cite{player_real-time_2023} are not filtered based on collision with the surrounding simulation environment, as the points input into the models are only points on the object.
To mitigate this, we transform the object's pose to make the grasp collision-free and kinematically feasible. 
If neither of these is achievable, the experiment is repeated.

\subsection{Real World Setup}
Real-world testing uses the underWater Arm-Vehicle Emulator (WAVE) at UW-APL \cite{rosette_wave_nodate}.  
It includes a Reach Robotics Bravo 7 manipulator \cite{noauthor_underwater_nodate} and a Trisect tri-focal underwater stereo sensor \cite{trisect_web_site} on a four-degree-of-freedom moving platform.
We test four scenarios of the kettlebell shown on right side in Fig. \ref{fig:hh101}, which also shows the test setup.
The platform was swept through various trajectories to allow the collection of multiple views of the target object.
We collect images of the kettlebell in challenging perceptual and grasping configurations: tilted towards the camera, lying flat on the ground, facing forward, and rotated sideways.
Images from the tests are shown on the bottom row of Fig. \ref{fig:qualitative_results}. 

We use RAFT-Stereo \cite{lipson2021raft} to compute depth from the stereo camera sensor and a fine-tuned Detectron2 \cite{wu2019detectron2} model to produce segmentation masks of the object.
The depth images and the left image from the stereo images are input into RTABMap \cite{labbe_rtabmap_2019} to estimate camera poses.
RTABMap produces poses with covariance estimates only at keyframes.
To increase the density of the reconstruction, we additionally use the poses between the keyframes and take the covariance from the most recent keyframe pose to use as the covariance for the pose estimates.

\section{Results and Discussion}
\subsection{Simulation Results}
We use three metrics to evaluate our method in simulation experiments.
First, we evaluate grasp traversal success, which indicates whether the end effector can move to the grasp pose without colliding with the object of interest.
When evaluating this, we set an initial pose with a 0.2 m offset from the proposed grasp pose.
We count a successful grasp traversal if no collisions occur when moving between the initial and final grasp poses.
Secondly, we evaluate whether the gripper can close without collision after reaching the proposed grasp pose. 
We count a success if the gripper successfully closes with part of the object inside the convex hull of the closed gripper.
Lastly, we evaluate whether, after closing the gripper, the arm can pick up the object 0.1 m above the position and hold it there for 5 seconds.

Table \ref{tab:quant_res_counting} shows the simulation experiments' results.
We point to the results in the partial and noisy reconstruction test setup.
TSGrasp consistently fails for partial reconstructions because the proposed grasp region is in an area with incomplete measurements.
This results in grasp candidates that consistently collide with the object.
Conversely, PUGS can determine that the edges of partially reconstructed geometry are unfavorable and guide the grasp to more reliable locations.
We show this occurring for the kettlebell and coffee mug in Fig. \ref{fig:sideways_fail}.
For the complete reconstruction, TSGrasp consistently outperforms PUGS.
This is expected, as the role of uncertainty as a signal to pull towards informative regions becomes less impactful when a complete reconstruction is obtained.

\subsection{Real World Results}
For the real-world results, we qualitatively evaluate the success of the grasp outputs for a kettlebell captured from different partial views.
Qualitative results from the grasp selection in real-world experiments are shown in Fig. \ref{fig:qualitative_results}.
We highlight that PUGS consistently leads the gripper pose to areas with more observations while retaining the grasping area's geometric feasibility.

The leftmost result in Fig. \ref{fig:qualitative_results} highlights the robustness that PUGS introduces when reasoning over very noisy pointclouds.
For this test, we tune the pointcloud filtering to be less aggressive by setting the outlier removal threshold to $\sigma_\text{thresh}=0.1$.
The proposed grasp pose from TSGrasp~\cite{player_real-time_2023} outputs the most confident grasp pose to be area captured due to noise, while PUGS successfully recovers to the kettlebell handle. 
This recovery of grasp success is shown with the green arrow in Fig. \ref{fig:qualitative_results} to indicate the correction.

\section{Conclusion and Future Work}
We present PUGS, an occupancy uncertainty estimation framework used for improving the robustness of grasp selection to measurement and pose uncertainty, which is common in underwater environments.
We perform experiments in simulation environments to quantify the quality of the proposed grasps from PUGS and showcase their improvement over baseline methods in challenging scenarios, such as noisy measurements or uninformative partial views. 
Real-world experiments are conducted in an underwater test tank to qualitatively evaluate the performance of the output of PUGS compared to baseline methods. Our results show that PUGS successfully leads the grasp location to ignore faulty and noisy measurements, with no hand-tuning required.

Future work will focus on modeling additional sources of uncertainty, such as measurement noise introduced from unfavorable visibility in underwater environments. 
Due to our formulation, additional uncertainties can be easily incorporated into PUGS. 
Currently, estimating the predictive uncertainty and sampling the occupancy field take on the order of minutes for the real world experiments. Optimization of similar methods has been performed for real-time probabilistic robotic perception~\cite{10093134}. This will be another direction for future research to enable the practical deployment of PUGS in the field.

\nocite{}
\bibliographystyle{IEEEtran}
\bibliography{ref}

\end{document}